\documentclass[10pt,twocolumn,letterpaper]{article}

\usepackage{iccv}              

\definecolor{iccvblue}{rgb}{0.21,0.49,0.74}
\usepackage[pagebackref,breaklinks,colorlinks,allcolors=iccvblue]{hyperref}

\usepackage{amsmath,amssymb}
\usepackage{graphicx}
\usepackage{caption}
\usepackage{subcaption}

\usepackage{booktabs}
\usepackage{float}
\usepackage{lipsum}
\usepackage{afterpage}
\usepackage{multirow}
\usepackage{colortbl}
\usepackage{hhline}
\usepackage{color, xcolor}
\usepackage{adjustbox}
\usepackage{tabularx}
\usepackage{siunitx}
\usepackage{pifont}
\usepackage{tikz}

\definecolor{darkred}{RGB}{150,0,0}
\definecolor{forestgreen}{RGB}{34,139,34}
\definecolor{mtlgreen}{RGB}{223,239,213}
\definecolor{mtlblue}{RGB}{213,223,241}
\definecolor{datablue}{RGB}{60,103,188}

\usepackage{hyperref}

\newcommand{\model}{GLOVER}
\newcommand{\age}{\texttt{AGE}}

\def\paperID{190} 
\def\confName{ICCV}
\def\confYear{2025}

\title{GLOVER: Generalizable Open-Vocabulary Affordance Reasoning \\for Task-Oriented Grasping}

\author{Teli Ma$^{1, \dagger}$, Zifan Wang$^{1, \dagger}$, Jiaming Zhou$^{1}$, Mengmeng Wang$^{2}$, Junwei Liang$^{1,3,*}$ \\ 
$^1$AI, HKUST(GZ) \quad 
$^2$ZJUT \quad  
$^3$CSE, HKUST \quad\\
{\tt\small tma184@connect.hkust-gz.edu.cn} \quad {\tt\small junweiliang@hkust-gz.edu.cn} \\
{\tt \small $\dagger$Equal Contribution} \quad {\tt \small *Corresponding Author} \\
\url{https://teleema.github.io/projects/GLOVER/}
}
\begin{document}

\twocolumn[{%
\renewcommand\twocolumn[1][]{#1}%
\maketitle


\begin{center}
    \centering
    \captionsetup{type=figure}
    \includegraphics[width=1.0\linewidth]{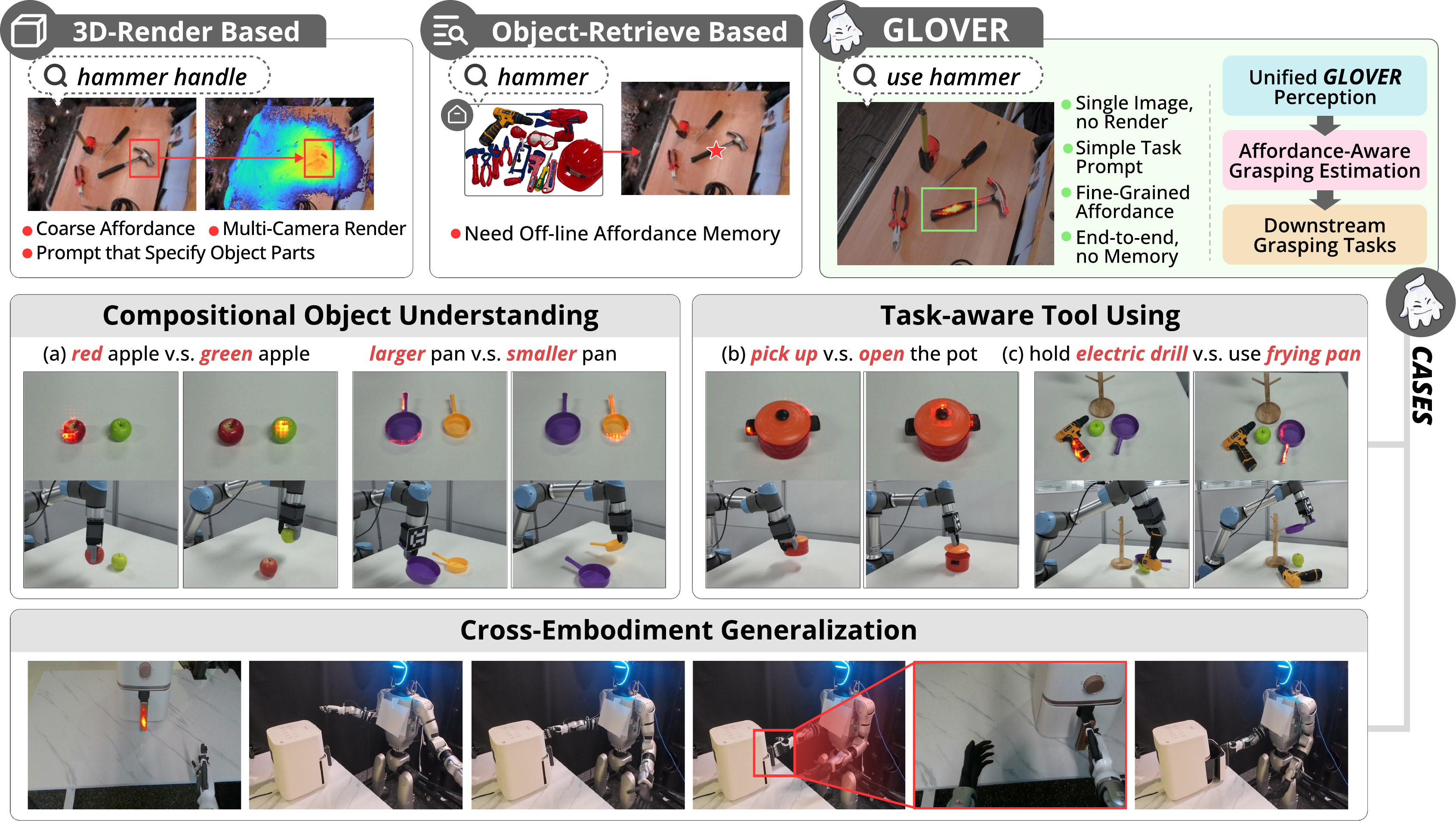}
    \vspace{-10pt}
    \caption{
    \textbf{Top:} Compared with previous methods, \model~eliminates the need for capturing multi-view images for 3D rendering, explicit instruction of the grasping part, and building extra off-line affordance memory. \model~is capable of providing more accurate fine-grained affordance predictions.   
    \textbf{Mid:} We demonstrate the efficacy of \model~in handling complex scenarios involving \textbf{compositional object understanding} and \textbf{task-aware tool using} under targeted human instructions, including: (a) attributes and relations like color, size, material, (b) tool function reasoning according to action, (c) common tool using in complex scenes. \textbf{Bottom:} We validate the effectiveness across embodiments (in a humanoid robot with dexterous hands), the robot grasps the oven handle based on inferred affordance.} 
    \label{fig:intro}
\end{center}%
}]

\maketitle
\begin{abstract}
Inferring affordable (i.e., graspable) parts of arbitrary objects based on human specifications is essential for robots advancing toward open-vocabulary manipulation. Current grasp planners, however, are hindered by limited vision-language comprehension and time-consuming 3D radiance modeling, restricting real-time, open-vocabulary interactions with objects. To address these limitations, we propose GLOVER, a unified Generalizable Open-Vocabulary Affordance Reasoning framework, which fine-tunes the Large Language Models (LLMs) to predict the visual affordance of graspable object parts within RGB feature space. 
We compile a dataset of over 10,000 images from human-object interactions, annotated with unified visual and linguistic affordance labels, to enable multi-modal fine-tuning. 
GLOVER inherits world knowledge and common-sense reasoning from LLMs, facilitating more fine-grained object understanding and sophisticated tool-use reasoning. 
To enable effective real-world deployment,  
we present Affordance-Aware Grasping Estimation (AGE), a non-parametric grasp planner that aligns the gripper pose with a superquadric surface derived from affordance data.
In evaluations across 30 table-top real-world scenes, GLOVER achieves success rates of 86.0\% in part identification and 76.3\% in grasping, with speeds approximately 29 times faster in affordance reasoning and 40 times faster in grasping pose estimation than the previous state-of-the-art. We also validate the generalization across embodiments, showing effectiveness in humanoid robots with dexterous hands.
\end{abstract}    
\vspace{-10pt}
\section{Introduction}

Human beings have an inherent ability to manipulate objects by understanding natural language instructions, such as distinguishing between different object types, identifying objects’ locations, and determining which part to grasp based on the desired task.
Motivated by this, research in robotic grasping has evolved from focusing on closed-set objects~\cite{peract, goyal2023rvt, ma2024contrastive} to open-vocabulary methods~\cite{kokic2020learning, lerftogo, zheng2024gaussiangrasper, geff, roboabc, okrobot, padalkar2023open, huang2024a3vlm, huang2023voxposer}. The previous methods mainly consisted of two categories, based on 3D radiance modeling~\cite{zheng2024gaussiangrasper, lerftogo, huang2023voxposer, geff} and object retrieving~\cite{kuang2024ram, roboabc} (the top row of Fig.~\ref{fig:intro}).
Methods based on 3D radiance modeling necessitate the acquisition of images from multiple cameras and the rendering of each scene, which is time-consuming. On the other hand, approaches based on object retrieving require the establishment of an additional offline object memory, which is labor-intensive.
Also, these methods suffer from a lack of complex reasoning capabilities regarding object properties. 
They face challenges in locating objects with \textbf{subtle linguistic distinctions} (e.g., distinguishing between a \textit{red apple} and a \textit{green apple} as shown in Fig.~\ref{fig:intro}) and determining \textbf{task-specific graspable parts} (e.g., \textit{pick up the pot} versus \textit{open the pot} as shown in Fig.~\ref{fig:intro}).

In this work,
we present a unified \underline{G}enera\underline{l}izable \underline{O}pen-\underline{V}ocabulary Affordanc\underline{e} \underline{R}easoning (\model) framework for open-vocabulary robotic grasping in an end-to-end manner. 
We aim to leverage the open-vocabulary reasoning capabilities of Large Language Models (LLMs) and fine-tune LLMs to output visual affordance masks.
To achieve this, we define the graspable region as a global \textbf{visual affordance mask},
inspired by the visual affordance inferring~\cite{hou2021affordance, luo2022learning, affcorrs, affordancellm} and reasoning segmentation task~\cite{lai2024lisa, groundingdino, rasheed2024glamm}. 
Unlike a binary mask, this affordance mask encodes a continuous probability map, representing the likelihood of grasping at various locations.
We adopt the affordance mask representation for two reasons: 
(1) graspable parts are better represented as regions rather than single points,
and (2) predicting global masks based on language input aligns more naturally with model decoding~\cite{llava, sam, lai2024lisa, he2022masked}.
To support this approach, we collect over 10,000 human-object interaction images, using Vision-Language Models (VLMs) to annotate the language labels and unified Gaussian distribution to annotate visual affordance masks.

With the collected dataset, 
we take the pairs of image and language instruction as input, leveraging LLaVA-7B~\cite{llava} to perform multi-modal encoding following~\cite{lai2024lisa}. 
The encoded affordance token, which aggregates both visual and linguistic features, is fed into an affordance decoder to output the desired affordance mask. 
This mask is then projected into 3D space as \textbf{stereo affordance} for downstream tasks.
The fine-tuning pipeline of \model~offers two key benefits:
(1) It can leverage extensive 2D human-object interaction images from diverse kinds of datasets~\cite{grauman2022ego4d, epickitchen, 3doi, agd20k}, overcoming the scarcity of 3D affordance data. 
(2) The fine-tuning allows \model~to inherit world knowledge and common-sense reasoning from the base LLM, enabling compositional object understanding and task-aware tool using in an open-vocabulary manner, as shown in Fig.~\ref{fig:intro}. 

To enable real-world grasping,
we introduce an \underline{A}ffordance-Aware \underline{G}rasping \underline{E}stimation (\age) module that estimates gripper poses based on the geometry of affordance regions.
\age~is a non-parametric method that outperforms learning-based grasping planners~\cite{graspnet, anygrasp} in both performance and efficiency (40 times faster), without reliance on additional training data.  
Inspired by~\cite{liu2022robust, vezzani2017grasping}, 
\age~samples grasping poses within the affordance space, determining the target pose by aligning the gripper with the superquadric surface derived from the stereo affordance geometry The alignment is optimized via nonlinear constrained optimization. 
This approach eliminates the usage of standalone grasping pose models~\cite{kuang2024ram,roboabc}, enabling direct and affordance-aware pose estimation in an end-to-end manner.

To summarize, our contributions are as follows:
(\textbf{i}) We present \model, a unified end-to-end perception framework designed for open-vocabulary robotic grasping. To enhance its capabilities, we have curated and annotated a dataset of over 10,000 multi-modal affordance images for fine-tuning, enabling \model~to leverage world knowledge and common-sense reasoning inherited from large language models (LLMs). (\textbf{ii}) We propose an \age~module to perform non-parametric grasping pose estimation, which demonstrates a \textbf{$\times$40} speed improvement over previous approaches. (\textbf{iii}) Our \model~module achieves state-of-the-art performance in the affordance benchmark, surpassing previous methods by a significant margin. We test the open-vocabulary grasping capability across 30 challenging table-top real-world scenes, demonstrating an average improvement of $\textbf{20.0\%}$ in affordance reasoning success rate and $\textbf{17.3\%}$ in grasping success rate compared to the previous state-of-the-art. We also demonstrate the generalization across diverse scenes and embodiments, showing effectiveness in 4 tasks with humanoid robots and dexterous hands.

\section{Related Work}

\subsection{Open-Vocabulary Representation for Manipulation}
Many recent studies have worked on language-guided robotic tasks like navigation~\cite{anderson2018vision, krantz2021waypoint, krantz2020beyond} and manipulation~\cite{di2024dinobot,okrobot, lerftogo}. The intervention of language plays a positive role in policy learning~\cite{goyal2023rvt, ma2024contrastive, huang2023voxposer}, value functions estimation~\cite{ahn2022can, liang2023code} and visual perception~\cite{geff, zheng2024gaussiangrasper}. Among them, the open-vocabulary manipulation is one of the most significant research topics. 
Several recent works focus on integrating 2D foundation models with 3D feature fields to achieve a 3D semantic-aware representation for open-vocabulary tasks~\cite{geff, lerftogo, shen2023distilled, zheng2024gaussiangrasper}. 
These methods distill features from 2D foundation models like CLIP~\cite{clip}, DINO~\cite{dino} as training objective for NeRF~\cite{nerf} or GaussianSplatting~\cite{gaussiansplat} to reconstruct 3D feature fields.  
In this work, we directly finetune the 2D foundation models to generate the related affordable areas in a generalizable and open-vocabulary manner, which is robust to scenario changes.

\subsection{Task-Oriented Grasping}
Task-oriented grasping refers to grasping different parts of objects based on the tasks. Previous research solves the task via detecting related object parts~\cite{myers2015affordance, chu2019learning, lu2023vl, huang2024a3vlm, li2024learning}, modeling 3D point clouds for affordance grounding~\cite{song2023learning, geng2023gapartnet, ling2024articulated}, or transferring grasps to new instances based on category~\cite{geng2023gapartnet, roboabc}.  
Recent works~\cite{huang2024a3vlm, di2024dinobot, song2023learning, lu2023vl, lerftogo, roboabc, shapegrasp} leverage vision-language models to reason the object parts for grasping. LERF-TOGO~\cite{lerftogo} derives a rough 3D object mask using DINO~\cite{dino} features to expand a relevant area locally. Subsequently, a LERF~\cite{lerf} query is conditioned on this mask to separate sub-parts of the object. 
Robo-ABC~\cite{roboabc} reasons the objects' grasping point by using CLIP~\cite{clip} to retrieve objects that share semantic similarity from the affordance memory. 
ShapeGrasp~\cite{shapegrasp} infers the contact points by prompting the large language models via Chain of Thought~\cite{wei2022chain}. 
Almost all the above methods rely on labeled 3D part-affordance datasets or additional pre-trained grasp networks like GraspNet~\cite{graspnet}, AnyGrasp~\cite{anygrasp} to infer grasping pose. However, \model~can leverage extensive 2D affordance data, well-trained 2D foundation models, and does not require an additional grasp planner network to estimate poses, which is more efficient.

\subsection{Visual Affordance Reasoning}
Previous research infers the affordance from human-object interactions~\cite{hassan2016attribute, hou2021affordance, luo2022learning}, scene understanding~\cite{lerftogo, lu2024manigaussian, geff} and 3D point cloud grounding~\cite{geng2023rlafford, ning2024where2explore, wu2024learning, mo2021where2act}. 
Recently, the foundation models such as LLMs and VLMs, have been integrated to perform affordance reasoning~\cite{affcorrs, affordancellm, roboabc, song2023learning, lu2023vl, xu2023joint}. 
 AffCorrs~\cite{affcorrs} tackle one-shot visual affordance transfer by querying the object parts to find semantically corresponding ones via pre-trained DINO-ViT~\cite{dino}.  AffordanceLLM~\cite{affordancellm} train the LLaVA~\cite{llava} on affordance dataset AGD20K~\cite{agd20k}, leveraging the world knowledge of the foundation model. Both Robo-ABC~\cite{roboabc} and RAM~\cite{kuang2024ram} adopt the retrieve-and-transfer framework for zero-shot affordance reasoning. They construct the affordance memory from 2D images and retrieve the similar demonstration from the affordance memory with the help of CLIP~\cite{clip} to reason affordance in the unseen domain. Our \model~does not require the creation of affordance memory, instead, it reasons the affordance leveraging LLM's world knowledge in an end-to-end manner.
 
\begin{figure*}[h!]
\centering
\includegraphics[width=1.0\linewidth,trim={0cm 0cm 0cm 0cm}]{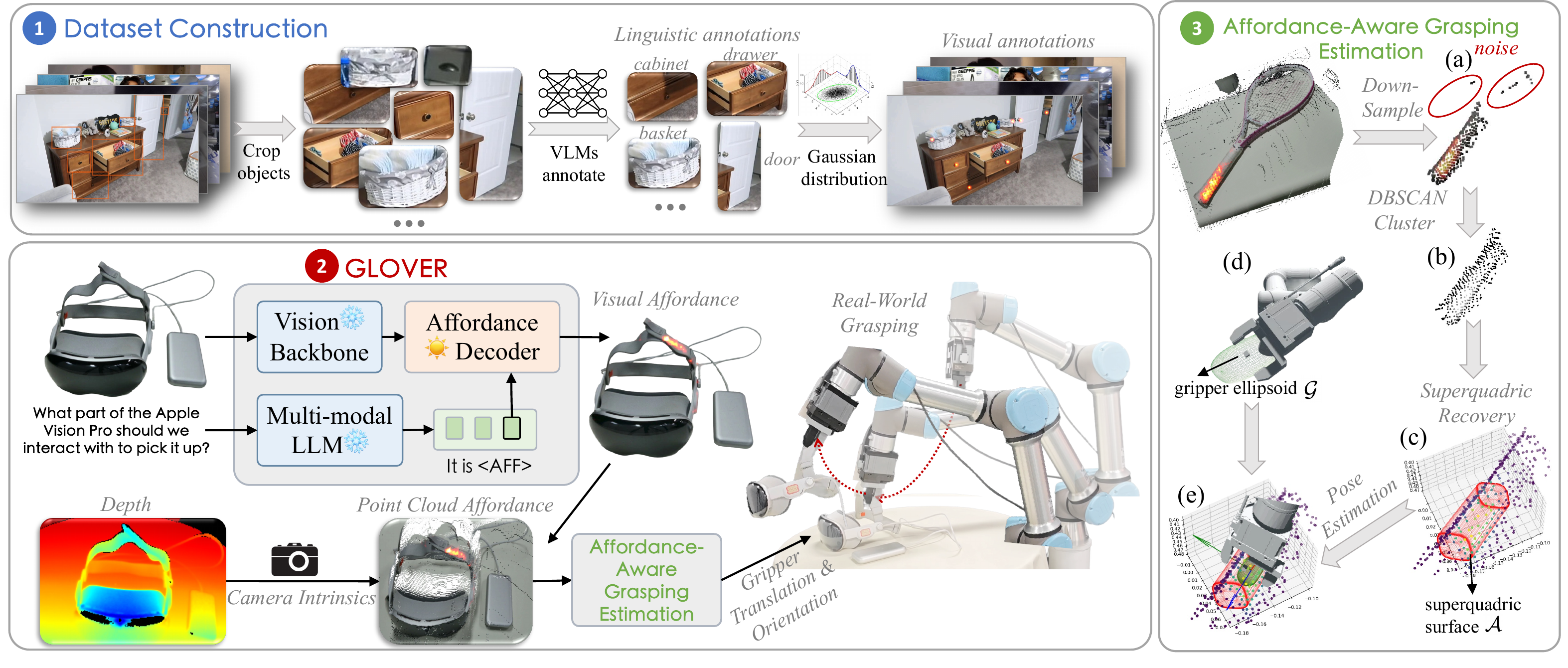}
\vspace{-20pt}
\caption{\textbf{An overview of our method} (Our contributions are highlighted with colored numbers). \textcolor{datablue}{1. We annotate the categories with the VLM and unify the affordance representation.} \textcolor{darkred}{2. We fine-tune the affordance decoder to decode the affordance token \texttt{[AFF]}, which encodes multi-modal information from multi-modal LLM. The fine-tuned \model~infers visual affordance in an open-vocabulary manner.} \textcolor{forestgreen}{3. The affordance-aware grasping estimation module (\age), including (a) Voxel down-sampled point clouds. (b) Filter the noise with DBSCAN~\cite{schubert2017dbscan} clustering. (c) Recover superquadric $\mathcal{A}$ from filtered stereo affordance. (d) Denote the gripper similarly as an ellipsoid surface $\mathcal{G}$. (e) Estimate the grasp pose by aligning the $\mathcal{A}$ and $\mathcal{G}$.}} 
\label{fig:pipeline}
\vspace{-15pt}
\end{figure*}

\section{GLOVER Method}
In this section, we tackle two key problems for open-vocabulary affordance reasoning. 
First, we identify an effective approach to represent visual affordance for fine-tuning the foundational model (Sec.~\ref{sec:data}).
Second, we examine how to integrate affordance knowledge into an existing foundational model while preserving as much of its original world knowledge as possible (Sec.\ref{sec:3.2}-~\ref{sec:3.4}).

\subsection{VL-Affordance Dataset Construction}
\label{sec:data}
\noindent \textbf{Image Data collection.} 
To enable affordance reasoning with the world knowledge of LLMs,
we leverage the abundant resources of 2D images with visual affordance annotations. 
We select images from two common affordance datasets, AGD20K~\cite{agd20k} and 3DOI~\cite{3doi}. 
AGD20K is a semi-supervised affordance dataset that includes 23,816 images with 50 categories of objects and 36 categories of affordance, sourced mainly from COCO~\cite{lin2014coco}, HICO~\cite{chao2015hico} and free-license websites.

For 3DOI, we use 10,000 images drawn from Articulation~\cite{articulation}, EpicKitchen~\cite{epickitchen}, and Taskonomy~\cite{zamir2018taskonomy}, which contain over 5,000 affordance-related interaction points. However, 3DOI does not include text labels for the interactable objects. 
Both AGD20K and 3DOI images feature a mix of egocentric and exocentric views.
We collect the linguistic and visual affordance annotations as follows.

\noindent\textbf{Generate the linguistic affordance labels.}
To address missing object labels in the 3DOI~\cite{3doi} dataset, we utilize VLMs for linguistic annotating, followed by human cross-validation to correct any errors. 
Specifically, we first crop objects from the background based on the bounding box annotations of the images. The cropped object images are then fed into a vision-language foundation model, SEED-X-17B~\cite{ge2024seed}, and prompt it with \textit{``What is the object in the image?"}. 
We manually filter the generated answers to retain only relevant object categories to ensure accurate annotations.
See dataset construction in Fig.~\ref{fig:pipeline}.

\noindent \textbf{Unify the visual affordance annotations.}
Visual affordances represent the interaction points where humans engage with objects. 
Predicting affordance through global masks from language inputs aligns more naturally with model decoding. 
Hence, following the approach in 3DOI~\cite{3doi}, we transform affordance points into a 2D Gaussian bump representation. 
For human-object interaction points in a 2D image, represented as $a_k = [x_k, y_k]$ for $k = 1, 2, \ldots, N$, we define the affordance probability mask as:
\begin{equation}\label{eq:gaussian}
\boldsymbol{\hat{M}}_{aff}(i, j) = {\rm exp}\left(-\frac{(i-x_k)^2 + (j-y_k)^2}{2\sigma^2}\right),
\end{equation}
where $\sigma$ is the standard deviation, and $(i, j)\in [0,W]\times[0,H]$ is the spatial indice within the image.

\noindent\textbf{Dataset overview.}
In the end, we obtain a dataset consisting of 12,215 images with 52,240 instances of human-object interactions. The images cover indoor and outdoor scenes, as well as egocentric and exocentric views. 
Each instance is annotated with both visual and linguistic affordance labels. We believe that our dataset will benefit future research in vision-language affordances.


\subsection{From Segmentation to Affordance Reasoning}
\label{sec:3.2} 
A key distinction between 2D affordance reasoning and traditional 2D instance segmentation is that affordance reasoning produces a continuous probability map, rather than a binary mask.
An intuitive idea is to extend VLMs for open-vocabulary segmentation, enabling them to perform open-vocabulary affordance reasoning. 
This approach enables affordance reasoning while preserving the foundational model’s open-world knowledge in a cost-effective way.

LISA~\cite{lai2024lisa} is a large language-instructed segmentation VLM built on an LLM. We adpot the LISA's structure, which includes a vision backbone, a multi-modal LLM (i.e., LLaVA~\cite{llava}), and a segmentation decoder. We initialize \model~with the LISA-7B pre-trained weights.

\subsection{Model Details}
\label{sec:3.3}
    \noindent\textbf{Prompt constructing.} 
    To guide the multi-modal LLM (LLaVA~\cite{llava}) in generating tokens for affordance decoding, we construct prompts in the format: \textit{``\texttt{<IMG>} What part of the \texttt{[OBJ]} should we interact with to \texttt{[ACT]} it?"}, where \texttt{<IMG>} represents image tokens, and \texttt{[OBJ]} and \texttt{[ACT]} specify the object name and action, respectively. 
    \textit{``\texttt{[ACT]} it''} will be removed from the prompt if the annotation does not exist.
    
    \noindent \textbf{Multi-modal encoding.}
    We follow the \textit{Embedding-as-Mask} paradigm in LISA, adding a new affordance token \texttt{<AFF>} to encode combined visual and linguistic features. Given a text prompt $\boldsymbol{t}$ and an input image $\boldsymbol{i}$, we feed them into the LLaVA model $\mathcal{F}_{LLM}$ to obtain a response $\boldsymbol{r}$ (of a sequence of feature vectors):
    \begin{equation}
        \boldsymbol{r} = \mathcal{F}_{LLM} (\boldsymbol{i}, \boldsymbol{t}),
    \end{equation}
    where the encoded $\texttt{<AFF>}$ token feature $\boldsymbol{u}$ is included in $\boldsymbol{r}$.
    
    \noindent \textbf{Visual encoding.}
    The visual features are important for affordance perception. We adopt the ViT~\cite{dosovitskiy2020vit} backbone $\mathcal{F}_{enc}$ to aggregate visual features, encoded as:
    \begin{equation}
        \boldsymbol{f} = \mathcal{F}_{enc}(\boldsymbol{i}),
    \end{equation}

    \noindent \textbf{Visual affordance decoding.}
    With the affordance token $\boldsymbol{u}$ carrying vision-language knowledge, we decode the visual affordance conditioned on it in the visual feature space $\boldsymbol{f}$.
    We follow LISA~\cite{lai2024lisa} and SAM~\cite{sam} to stack Transformer decoder blocks for affordance decoding.  
    Each decoder block consists of self-attention and bi-directional cross-attention. This process is formulated as:
    \begin{equation}
        \boldsymbol{M}_{aff} = \mathcal{F}_{dec}(\boldsymbol{u}, \boldsymbol{f}),
    \end{equation}
    where $\mathcal{F}_{dec}$ represents the visual affordance decoder. Please refer to Fig.~\ref{fig:pipeline} for the pipeline.

\subsection{Training Objective}
\label{sec:3.4}
To preserve the foundational model's world knowledge, we freeze the language model component and fine-tune only the affordance decoder parameters specific to our task. 
Hence, \model~is trained end-to-end with only affordance loss to update the affordance decoder.
Unlike segmentation mask decoding, which often relies on cross-entropy and DICE losses, we employ the sigmoid focal loss~\cite{lin2017focal} for affordance decoding, as it better handles the continuous distribution of affordance probabilities. The training objective is defined as:
\begin{equation}
    \mathbb{L} = \mathcal{L}_{aff} = \mathbf{FL}(\boldsymbol{M}_{aff}, \boldsymbol{\hat{M}}_{aff}).
\end{equation}

\section{Affordance-Aware Grasping Estimation}
\label{sec:4}
With the inferred visual affordance, the next question is determining how the agent can interact with objects based on these affordances to effectively manipulate them. This question is challenging as the projected stereo affordances (see Sec.\ref{sec:4.1}) from the visual affordances have complicated and irregular geometric surfaces, making it difficult to identify a global optimal point to grasp.
Superquadric recovery~\cite{liu2022robust, vezzani2017grasping, solina1990recovery, leonardis1997superquadrics, chevalier2003segmentation, paschalidou2019superquadrics, paschalidou2020learning} is an effective method for estimating superquadric of irregular surfaces, which can be used to estimate the geometry of stereo affordance. Based on the estimated superquadric of affordance region, the grasping pose is calculated via nonlinear constrained optimization. We describe the details below (and illustrated in the third part of Fig.~\ref{fig:pipeline}).

\subsection{Stereo Affordance Preprocessing}
\label{sec:4.1}
We first map the deduced visual affordance onto stereo space using the RGB-D camera's intrinsic parameters. 
Since the model may infer multiple clusters in the stereo affordance space, we filter out those with low affordance weights.
Specifically, we first down-sample the point cloud with voxel down-sampling.
Then, we apply DBSCAN~\cite{schubert2017dbscan}, a density-based clustering algorithm, to divide the affordances into distinct clusters. 
We calculate the mean affordance weight for each cluster and discard the 3D point clouds of clusters with low mean weights.

\subsection{Superquadric Recovery from Stereo Affordance}
Superquadrics represent a class of geometric shapes that can model diverse forms. To determine the grasping pose from complex affordance regions, we reconstruct superquadrics from the geometric structures of the affordance point clouds, representing each superquadric as $\mathcal{A}$. 
This process centers on identifying an optimal set of parameters $\boldsymbol{\lambda} \in \mathbb{R}^{11}$ to maximize alignment between $N$ stereo affordance points $\boldsymbol{a}_i = [x_i, y_i, z_i]$ ($i = 1, \ldots, N$) and the superquadric surface.

Inspired by superquadric recovery methods~\cite{liu2022robust, vezzani2017grasping}, we minimize the radial Euclidean distance from the $\boldsymbol{a}_i$ to the superquadric surface $\mathcal{A}$ for aligning. 
To integrate the affordance weight, this process can be formulated as:

\begin{equation}\label{eq:object}
\min _{\boldsymbol{\lambda_\mathcal{A}}} \sum_{i=1}^N\left(\mathcal{W}_i\sqrt{\lambda_V}\left(F\left(\boldsymbol{a}_i, \boldsymbol{\lambda_\mathcal{A}}\right)-1\right)\right)^2 + \beta V\left(\boldsymbol{\lambda_\mathcal{A}}\right).
\end{equation}
The affordance weight $\mathcal{W}_i$ ensures the inclination towards affordance points with high probability when estimating surface geometries of superquadrics.
The inside-outside function $\left(F\left(\boldsymbol{a}_i, \boldsymbol{\lambda_\mathcal{A}}\right)-1\right)^2$ aims to minimize the radial Euclidean distance, while the term $\lambda_V$ is the superquadric volume coefficient, deprecating the expansion of the superquadric volume. 
To further control the range of the recovered superquadric, we construct a penalty term based on the estimated volume of superquadrics, termed as $\beta V(\boldsymbol{\lambda_\mathcal{A}})$  to ensure the robustness of the method for noisy point clouds.

\subsection{Grasp Pose Estimation}

We denote the gripper similarly as an ellipsoid surface following previous works~\cite{vezzani2018improving,liu2022robust}. The gripper's pose is defined by a 7D vector $\boldsymbol{x}=\left[x_g, y_g, z_g, {q_g}^x,{q_g}^y,{q_g}^z,{q_g}^w\right]$, where $\left(x_g, y_g, z_g\right)$ are the coordinates of the gripper's position and $\left({q_g}^x,{q_g}^y,{q_g}^z,{q_g}^w\right)$ are its orientation quaternions. For poses $\boldsymbol{x}$, our goal is to identify a pose $\boldsymbol{\hat{x}}$ that allows the gripper ellipsoid $\mathcal{G}$ to align with the affordance superquadric $\mathcal{A}$ while satisfying constraints that ensure $\boldsymbol{\hat{x}}$ is within the gripper's reach (the third part in Fig.~\ref{fig:pipeline}).

This can be formalized as the following nonlinear constrained optimization problem:

\begin{equation} \label{eq:gripper}
\begin{aligned}
\boldsymbol{\hat{x}} = &\arg \min _{\boldsymbol{x}} \sum_{i=1}^L\left(\left(F\left(\boldsymbol{p}_i^{\boldsymbol{x}}, \boldsymbol{\lambda_{\mathcal{A}}}\right)-1\right)\right)^2, \\
& \text { subject to: } \\
& C_i\left(\boldsymbol{c}_i,\boldsymbol{p}_1^{\boldsymbol{x}}, \ldots, \boldsymbol{p}_L^{\boldsymbol{x}}\right)>0. 
&
\end{aligned}
\end{equation}
The cost function Eq.\eqref{eq:gripper} aims to minimize the distance between the affordance superquadric $\mathcal{A}$ and $L$ points $\boldsymbol{p}_i^{\boldsymbol{x}}$, while $\boldsymbol{p}_i^{\boldsymbol{x}}$ are sampled from the closest half of the gripper ellipsoid $\mathcal{G}$. This choice prevents the gripper from penetrating the object by ensuring that only the nearest portion of $\mathcal{G}$ approaches $\mathcal{A}$, thus avoiding potential collisions.~Constraint terms $C_i(\cdot)$ are employed to ensure the generation of safe grasping poses, such as avoiding self-collision and collision with enviromental obstacles.

\begin{table}[!htb]
\renewcommand{\arraystretch}{1.2}
\centering
\large
\resizebox{1.0\linewidth}{!}{
\begin{tabular}{c|c|c|c}
\hline
 \textbf{Task} & \textbf{Sub-Tasks} &\textbf{\#Num} & \textbf{Objects}  \\ 
 \hline
 &\multirow{2}*{Attribute} &\multirow{2}*{7} &apple, mug, electric drill,\\
\textbf{Compositional}& & &pan, cup, spoon, fork\\ \cline{2-4}
\textbf{Object}  &Relation &3 &mug, cup, pan\\ \cline{2-4}
 \textbf{Understanding} & Complex &\multirow{2}*{5} &tape, goggles, tennis racket,\\
 &Scene$^*$ & &electric drill, fruit\\ 
 \hline
 &\multirow{3}*{Tool Use} &\multirow{3}*{10} &screwdriver, knife, scissors, \\ 
  \textbf{Task-Aware} & & &hammer, pliers, electric drill,\\
  \textbf{Tool Using} & & &tape measure, saw, pot, pan\\ \cline{2-4}
  &Function & \multirow{2}*{5} &knife, charger, sanitizer\\
 &Reason & &bottle, pot, tissue box\\
\hline
\end{tabular}
}
\caption{\textbf{The specific details of testing scenes we used in the model evaluation.} 
\model~can conduct open-vocabulary affordance reasoning and grasping on nearly any object encountered in daily life. Balancing experimental requirements and practical constraints, we selected specific objects to construct the experiments, with the aim of standardizing the testing of different models' capabilities. (* In the Complex Scene, the distractors are from all objects we own, over 5 items per scene.)}
 \label{tab:objects}
 \vspace{-10pt}
\end{table}

\begin{figure*}[htbp!]
\centering
\includegraphics[width=1.0\linewidth,trim={0cm 0cm 0cm 0cm}]{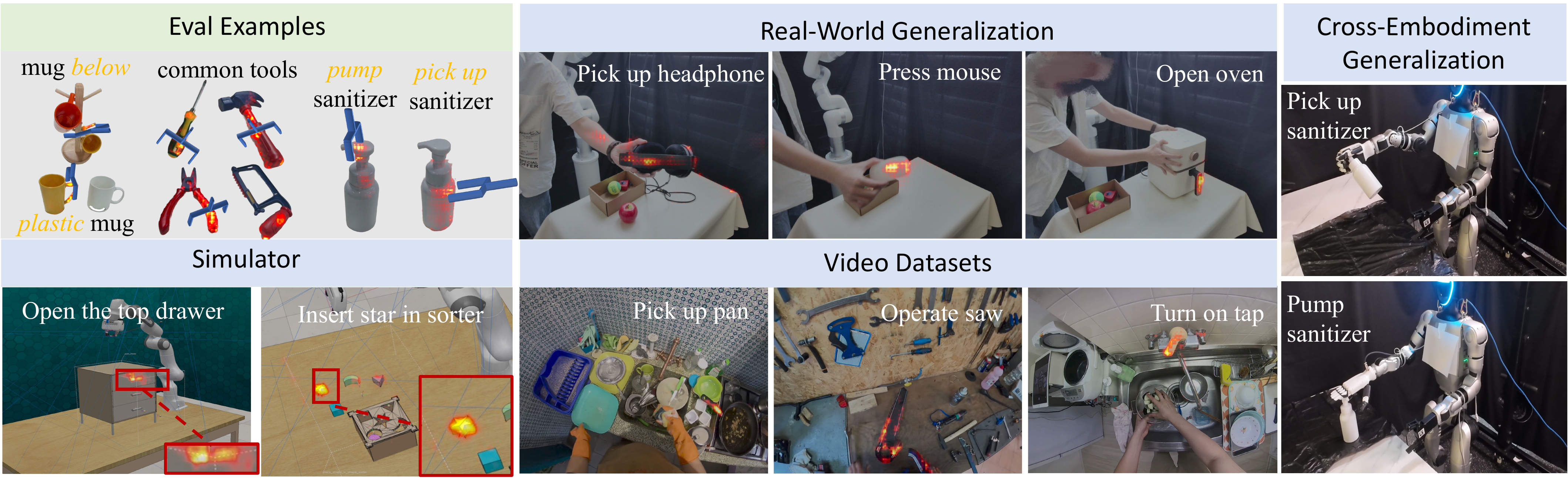}
\caption{\colorbox{mtlgreen}{\textbf{Eval Examples:}} Examples of inferred visual affordance and grasping pose in multiple scenes. The testing scenes are designed to evaluate the model's compositional object understanding (attributes, relations, complex scenes) and task-aware tools using (tool using, function reasoning). \colorbox{mtlblue}{\textbf{Generalization:}} \model~presents open-vocabulary ability across diverse environments, including real-world, simulator (RLBench~\cite{james2020rlbench}), scenes from other datasets (Ego4D~\cite{grauman2022ego4d}), and across-embodiments (humanoid robots with dexterous hands).}
\label{fig:data}
\end{figure*}

\begin{table*}[htbp!]
\renewcommand{\arraystretch}{1.2}
\centering
\resizebox{0.9\textwidth}{!}{
\begin{tabular}{@{}c|cccccc|cccc|cc@{}}
\hline
& \multicolumn{6}{c|}{\textbf{Compositional Object Understanding (\%)}}  & \multicolumn{4}{c|}{\textbf{Task-Aware Tool Using (\%)}} &\\ \cline{2-11}  
 \textbf{Method} & \multicolumn{2}{c}{Attribute(70)} & \multicolumn{2}{c}{Relation(30)} & \multicolumn{2}{c|}{Complex Scene(50)} & \multicolumn{2}{c}{Tool Use(100)} &\multicolumn{2}{c|}{Function Reason(50)} &\multicolumn{2}{c}{\textbf{\#AVG (\%)}} \\  
  & \texttt{Aff.}  &\texttt{Real}  &\texttt{Aff.}  &\texttt{Real} &\texttt{Aff.}  &\texttt{Real} &\texttt{Aff.}  &\texttt{Real} &\texttt{Aff.}  &\texttt{Real} &\texttt{Aff.}  &\texttt{Real} \\
 \hline
 VRB~\cite{vrb} &52.9 & 37.1 &36.7 &20.0 &8.0 &6.0 &58.0 &32.0 &0.0 &0.0 &36.7 &22.3\\
 LERF-TOGO~\cite{lerftogo} &58.6 &45.7 &30.0 &23.3 &76.0 &62.0 &64.0 &53.0 &22.0 &10.0 &54.7 &42.7 \\
 LISA$^*$~\cite{lai2024lisa} &- &34.0 &- &13.3 &- &14.0 &- &7.0 &- &0.0 &- &10.7 \\
 RAM~\cite{kuang2024ram} &64.3 &57.1 &43.3 &26.7 &78.0 &62.0 &84.0 &62.0  &34.0 &12.0 &66.0 &49.0 \\
 \rowcolor{gray!20}
 \textbf{\model} &\textbf{95.7} &\textbf{84.3} &\textbf{80.0} &\textbf{73.3} &\textbf{82.0} &\textbf{68.0} &\textbf{88.0} &\textbf{80.0} &\textbf{76.0} &\textbf{68.0}  &\textbf{86.0} &\textbf{76.3}\\
\hline
\end{tabular}
 }
\caption{\textbf{Real-world experimental results.} The \textit{Attribute, Relation, Complex Scene, Tool Use} and \textit{Function Reason} have 7, 3, 5, 10, 5 test scenes, respectively. All the scenes are evaluated 10 times by varying the object arrangements. We report the affordance reasoning and real-world grasping success rates to compare with previous methods.}
\vspace{-10pt}
\label{tab:real}
\end{table*}

\section{Experiments}

In this section, we establish a comprehensive experimental framework to validate the effectiveness and efficiency of \model. 
We compare with previous methods in both the real-world grasping (Sec.~\ref{sec:real-world-results}) and visual affordance reasoning benchmark (Sec.~\ref{sec:affordance-results}). Ablations (Sec.~\ref{sec:ablations}) are also performed to analyze the model components.

\subsection{Implementation Details}
We fine-tuned our \model~based on LISA-7B~\cite{lai2024lisa}, LLaVA~\cite{llava}, and SAM~\cite{sam} on eight NVIDIA A6000 GPUs for $5$ epochs, requiring approximately 18 hours to complete. The initial learning rate is $5e-5$ and we use the AdamW~\cite{adamw} optimizer by default. 
For the real-world experiments, we collect the RGB-D images with an Orbbec Femto Bolt, with an image size of $1280\times 960$ for UR5e robot arm. The humanoid robot, Unitree G1, is utilized for cross-embodiment generalization with more real downstream experiments. The G1 robot is equipped with Inspire dexterous hands RH56DFX, and we use the original Intel RealSense D435i of the robot to capture RGB-D images with the size of $640\times480$. The details can be found in supplementary material.

\subsection{Real-World Task Design}
\label{sec:real-world}
We evaluate \model~on a wide range of objects from diverse scenes, 
focusing on two main challenges: \textit{compositional object understanding}, and \textit{task-aware tool using}. The scenes and objects we used are elaborated in the Table~\ref{tab:objects}.

\noindent\textbf{Compositional object understanding} refers to the agent's understanding of the fine-grained object attributes like colors, sizes, materials and relations in a compositional way.  
Here, the agent must select the correct object based on specific human instructions, even when the instructions are morphologically similar. 
This task is highly challenging, as it requires the agent to correctly interpret affordances associated with each object under potentially confusing instructions.

We design three sub-tasks for compositional object understanding, namely \textit{attributes}, \textit{relations}, and \textit{complex scenes}. The \textit{attributes} task evaluates the agents' understanding of colors, sizes, and materials of different objects. The \textit{relations} task challenges the agents' ability to understand spatial relations, like ``above" or ``below" mug shown in Fig.~\ref{fig:data}. For \textit{complex scenes}, we assess the model’s capacity for object-level perception by increasing the difficulty for the models to select objects based on language instructions through complex multi-object scenes.

\noindent\textbf{Task-aware tool using} 
requires reasoning about which parts of common tools are relevant to various everyday tasks based on vague functional descriptions. 
We split the task into two sub-tasks, namely \textit{tool use} and \textit{function reason}.

For \textit{tool use}, we prompt \model~ with phrases composed of verbs plus nouns, such as \textit{“pick up the hammer”}, to assess the agent’s understanding of the general affordances of everyday tools. 
In contrast, \textit{function reasoning} tests the agent’s understanding of the functions associated with different parts of a tool. 
This task requires more precise visual perception and sophisticated reasoning, as different verbs applied to the same tool (e.g., ``pick up'' vs. ``pump'' the sanitizer) imply varying affordances. This setup demands intricate common-sense reasoning from the agent, as shown in Fig.~\ref{fig:data}.

\noindent\textbf{Evaluation.} 
For each scene, we vary the object positions and orientations to test it 10 times, and report the success rate. The success rate includes affordance and grasping success rate, referring to the success rate of affordance reasoning and real-world grasping. 
We manually defined the ground truth regions for different objects based on the language instructions of the required task. As long as \textbf{the majority of} filtered affordance points fall within these ground truth regions, it is considered an affordance success. Grasping success is defined as the ability to complete real-world grasping.
For real-world grasping, we only test the scenes with successfully reasoned affordance.

\subsection{Real-World Results}
\label{sec:real-world-results}

\textbf{Baselines.}
We construct four baselines for comparisons. VRB~\cite{vrb} predicts contact points by learning from human video demonstrations. 
LERF-TOGO~\cite{lerftogo} reconstructs the scenes dynamically via LERF~\cite{lerf}, extracting 3D object masks from DINO~\cite{dino} features and conditioning object-part queries based on these masks. We also replace our \model~module with the original LISA-7B~\cite{lai2024lisa} model, which we denote as LISA$^*$, to highlight the effectiveness of our affordance fine-tuning.
RAM~\cite{kuang2024ram} constructs the affordance memory from 2D images and retrieves similar demonstrations from the affordance memory with the help of CLIP~\cite{clip} to reason affordance in the unseen domain.
For a fair comparison, we provide LERF-TOGO with ambiguous language queries rather than specific part queries (which require an additional LLM to infer).
Since LISA$^*$ outputs binary masks, we estimate grasping poses using superquadrics based on the stereo binary mask and report the real-world grasping results.

\noindent\textbf{Results.} The performance are reported in Table~\ref{tab:real}. 
Our model surpasses the previous approach, RAM~\cite{kuang2024ram}, achieving an average increase of 20.0\% in affordance success rate and 17.3\% in real-world grasping success rate.  
Key findings include: (1) VRB~\cite{vrb} performs poorly in scenes that need reasoning based on human instructions due to the lack of language processing capabilities; 
(2) LERF-TOGO performs well with object recognition but lacks the complex reasoning required for interpreting object relations and diverse tool functions, likely due to the bag-of-words~\cite{bagofwords} limitation;
(3) In contrast to the original LISA model, our model exhibits finer object component perception, resulting in more precise interactive regions and improved real-world grasping accuracy.
In summary, \model~excels in both intricate common-sense reasoning and precise affordance inference.

\subsection{Affordance Comparisons}
\label{sec:affordance-results}
We follow previous SOTA methods to compare affordance capabilities in the following benchmark. AGD20K~\cite{agd20k} is a large-scale affordance dataset with a test split for fair comparisons. We follow AffordanceLLM~\cite{affordancellm} and LOCATE~\cite{li2023locate} to report the results evaluated on the hard split of AGD20K testing. 

\noindent\textbf{Evaluation metrics.}
Following the previous work, we adopt the Kullback-Leibler Divergence (KLD), Similarity (SIM), and Normalized Scanpath Saliency (NSS) as evaluation metrics. 
Lower KLD values and higher SIM and NSS values indicate better affordance inference.
Details on these metrics are elaborated in the supplementary material.

\begin{table}[htbp!]
\renewcommand{\arraystretch}{1.0}
   \centering
   \vspace{-0.6em}
   \scalebox{0.9}{
   \begin{tabular}{l|lccc}
      \hline
      \textbf{Methods} & \textbf{KLD} $\downarrow$ & \textbf{SIM} $\uparrow$ & \textbf{NSS} $\uparrow$ \\
      \hline
      Cross-View-AG~\cite{luo2022learning}  & 2.092 & 0.209 & 0.138    \\
      Cross-View-AG+~\cite{luo2024grounded}  & 2.034 & 0.218 & 0.342   \\
    LOCATE~\cite{li2023locate}  & 1.829 & 0.282 & 0.276 \\
    LOCATE-Sup~\cite{li2023locate} &  2.003 & 0.224 & 0.435  \\
    LOCATE-Sup-OWL~\cite{li2023locate,minderer2022simple}   & 2.127& 0.206 & 0.314   \\
      3DOI~\cite{3doi} & 4.017 & 0.200 & 0.549 \\
      VRB~\cite{vrb} &2.154 &0.258 &0.236\\
      AffordanceLLM~\cite{affordancellm}  &\underline{1.661} & \underline{0.361} &\underline{0.947}\\
      \rowcolor{gray!20}
      \textbf{\model} & \textbf{1.098} & \textbf{0.476} & \textbf{1.552} \\
      \hline
   \end{tabular}
    }
   \caption{\textbf{Visual affordance results in the benchmarking dataset.} \model~outperforms previous state-of-the-art methods by a large margin in all three metrics.}
   \label{tab:affordance}
   \vspace{-10pt}
\end{table}

\noindent\textbf{Results.}
The results in Table~\ref{tab:affordance} show that \model~significantly outperforms previous methods across all three metrics.
Note that the testing images are \textbf{filtered out} from the training set of \model~in this comparison. 
The results highlight the effectiveness of our fine-tuning method in enhancing foundation models for open-vocabulary affordance reasoning.

\subsection{Ablations}
\textbf{Efficiency analysis.}
\label{sec:efficiency}
We assess \model’s efficiency in terms of time required for affordance reasoning and grasping pose estimation. 
The process of affordance reasoning includes both scene capture and model inference.  
As shown in Table~\ref{tab:time}, \model~achieves approximately 330 times and 29 times faster affordance reasoning compared to LERF-TOGO~\cite{lerftogo} and RAM~\cite{kuang2024ram}.
For grasping pose estimation, our proposed \age~(Sec.~\ref{sec:4}) is about 40 times faster than the GraspNet~\cite{graspnet} used in LERF-TOGO and RAM. 
All the time costs are reported on a single NVIDIA 4090 GPU.
The results demonstrate the efficiency of our approach. This high efficiency enables \model~to track moving objects dynamically, deducing affordances in a real-time way. The tracking performance can be found in the supplementary materials.

\begin{table}[htbp]
\centering
\small
\vspace{-8pt}
    \resizebox{0.88\columnwidth}{!}{
    \begin{tabular}{ccc}
        \toprule
       & \textit{Affordance Inference} &\textit{Grasping Pose} \\
       \multirow{-2}{*}{Methods} & \textit{Time} (\textit{s}) & \textit{Time} (\textit{s}) \\ \midrule
    LERF-TOGO~\cite{lerftogo} &$\sim$230.0 &$\sim$4.0  \\
    RAM~\cite{kuang2024ram}  &$\sim$20.0 &$\sim$4.0 \\
      \textbf{\model} &\textbf{$\sim$\textbf{0.7}}  & \textbf{$\sim$\textbf{0.1}}  \\
      \bottomrule    
    \end{tabular}}
    \vspace{-8pt}
    \caption{The time costs of affordance reasoning and grasping pose process.} 
    \label{tab:time}
    \vspace{-10pt}
\end{table}

\noindent\textbf{Grasping pose estimation module.}
\label{sec:ablations}
We ablate the grasping pose estimation module to show the effectiveness of the proposed \age. 
We evaluate the performance of its modules from two aspects. The first one is the real-world grasping success rate. 
Secondly, we assess the pixel-wise spatial distance (PWS-Distance) 
between the predicted grasp point and the point of maximum probability in the visual affordance map, as a direct measurement of grasping pose quality. 
We compare \age~with the popular GraspNet~\cite{graspnet} and AnyGrasp~\cite{anygrasp}, both of which are learning-based pose estimation models trained on large datasets. 
Each scene in Sec.~\ref{sec:real-world} is tested five times, and we report the average real-world success rate and PWS-Distance across all scenes.
Results in Table~\ref{tab:grasp} show that \age~outperforms both learning-based methods in both metrics.

\begin{table}[htbp!]
   \centering
   \small
   \vspace{-8pt}
    \resizebox{0.9\columnwidth}{!}{
   \begin{tabular}{ccc}
      \toprule
      Methods & \textit{Real-World SR.} (\%) $\uparrow$  & \textit{PWS-Distance} $\downarrow$ \\
      \midrule
      GraspNet~\cite{graspnet} &62.0 &0.179  \\
      AnyGrasp~\cite{anygrasp} &73.3 &0.183  \\
      \textbf{\age} &\textbf{78.7}  &\textbf{0.084}   \\
      \bottomrule
   \end{tabular}
    }
    \vspace{-8pt}
   \caption{\textbf{The performance of ablating different grasping pose estimation methods.} Real-World SR. and PWS-Distance represent real-world grasping success rate (\%) and pixel-wise spatial distance (between $[0,1]$), respectively.}
   \label{tab:grasp}
   \vspace{-10pt}
\end{table}

\noindent\textbf{Generalization.}
We show the visualization of open-vocabulary affordance reasoning in diverse scenarios to show the generalization in Fig.~\ref{fig:data}, including \textit{real-world, simulator, other video datasets}.
We also validate \model's generalization ability across embodiments, showing the effectiveness of the proposed method in humanoid robots with dexterous hands. The Inverse Kinematics is utilized to determine the motions for reaching the target grasping pose output by \model. We conduct experiments on four tasks, \textit{open oven}, \textit{pump sanitizer}, \textit{pick up sanitizer}, and \textit{pick up mug}. The results are shown in Table~\ref{tab:humanoid}. More details and failure cases are elaborated in the supplementary material.

\begin{table}[htbp]
\centering
\small
\vspace{-7pt}
    \resizebox{0.85\columnwidth}{!}{
    \begin{tabular}{ccccc}
        \toprule
      \multirow{2}*{Task} & \textit{Open}   & \textit{Pump} &  \textit{Pick up}  & \textit{Pick up} \\ 
      &Oven &Sanitizer &Sanitizer &Mug \\ \midrule
      \textbf{\model} &4/5 &3/5 &2/5 &3/5\\
      \bottomrule    
    \end{tabular}}
    \vspace{-8pt}
    \caption{Success rate of \model~in humanoid robot with dexterous hands.} 
    \label{tab:humanoid}
    \vspace{-10pt}
\end{table}

\section{Conclusion}
In conclusion, our \model~framework demonstrates substantial advancements in open-vocabulary affordance reasoning and task-oriented grasping, outperforming existing methods in both accuracy and efficiency. By fine-tuning a foundational model with enhanced affordance understanding while preserving world knowledge, \model~achieves state-of-the-art performance across real-world grasping and affordance reasoning tasks. The \age~module enables rapid and precise grasping pose estimation in a non-parametric manner. Extensive experiments and ablation studies confirm \model’s capabilities in complex reasoning, efficient affordance inference, and robust grasping across diverse scenarios and embodiments.

{
    \small
    \bibliographystyle{ieeenat_fullname}
    \bibliography{main}
}

\appendix
\section{Evaluation Metrics}
We elaborate on the three metrics we used in evaluating the \model~in the  affordance benchmark. 

\noindent\textbf{Kullback-Leibler Divergence (KLD)}  quantifies the distribution variance between the predicted affordance map $\boldsymbol{M}_{aff}$ and the ground truth $\boldsymbol{\hat{M}}_{aff}$ ( $\boldsymbol{M}_{aff}$, $\boldsymbol{\hat{M}}_{aff} \in \mathbb{R}^{H\times W}$). We first calculate the min-max normalization for each pixel in the $\boldsymbol{M}_{aff}$ and $\boldsymbol{\hat{M}}_{aff}$.
\begin{align}
    \boldsymbol{\hat{M}}_{aff}^{i} = \boldsymbol{\hat{M}}_{aff}^{i}&/\sum\boldsymbol{\hat{M}}_{aff}, \\
    \boldsymbol{M}_{aff}^{i} = \boldsymbol{M}^{i}_{aff}&/\sum \boldsymbol{M}_{aff}.
\end{align}
Then the KLD is formulated as:
\begin{equation}
     {\rm KLD}(\boldsymbol{\hat{M}}_{aff}||\boldsymbol{M}_{aff}) = \sum\limits_{i} \boldsymbol{\hat{M}}_{aff}^{i} \cdot log (\frac{\boldsymbol{\hat{M}}_{aff}^{i}}{\boldsymbol{M}^{i}_{aff}}).
\end{equation}

\noindent\textbf{Similiary (SIM)}, also known as histogram intersection, quantifies the overlap between the predicted affordance map $\boldsymbol{M}_{aff}$ and the ground truth $\boldsymbol{\hat{M}}_{aff}$.
\begin{equation}
    {\rm SIM}(\boldsymbol{M}_{aff}, \boldsymbol{\hat{M}}_{aff}) = \sum\limits_{i} min(\boldsymbol{M}_{aff}^{i}, \boldsymbol{\hat{M}}_{aff}^{i}).
\end{equation}

\noindent\textbf{Normalized Scanpath Saliency (NSS)} evaluates the alignment between the $\boldsymbol{M}_{aff}$ and the ground truth $\boldsymbol{\hat{M}}_{aff}$. We first pre-process the $\boldsymbol{M}_{aff}$ and $\boldsymbol{\hat{M}}_{aff}$ as:
\begin{equation}
    \mathcal{\hat{M}} = \mathbb{I} (\boldsymbol{\hat{M}}_{aff}>0.1),
\end{equation}
\begin{equation}
    \mathcal{M} = \frac{\boldsymbol{M}_{aff}-\mu(\boldsymbol{M}_{aff})}{\sigma(\boldsymbol{M}_{aff})},
\end{equation}
where $\mathbb{I}$ is the indicator function and $\mu, \sigma$ represent the mean and standard deviation of $\boldsymbol{M}_{aff}$.NSS is calculated as the mean of the normalized predictions at binary ground truth locations:

\begin{equation}
    {\rm NSS}(\mathcal{M}, \mathcal{\hat{M}}) = \frac{1}{\sum \mathcal{\hat{M}}}\sum\limits_{i}(\mathcal{M}\times \mathcal{\hat{M}}_{i}).
\end{equation}

\section{Real-World Experiment Settings}
We introduce the settings of our real-world experiments, including the single table-top robotic arm experiments and the humanoid robot experiments.

\begin{figure}[h!]
\centering
\includegraphics[width=0.98\linewidth,trim={0cm 0cm 0cm 0cm}]{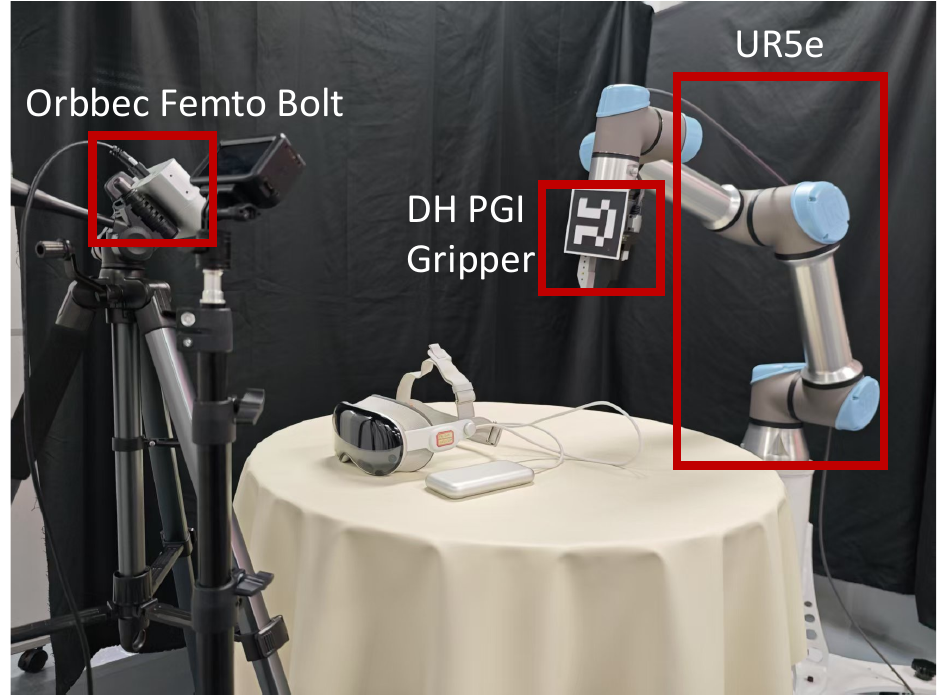}
\caption{The experiment setting of single table-top robotic arm.} 
\label{fig:ur5e}
\end{figure}

For the single table-top robotic arm, we utilize a UR5e robotic arm equipped with a DH PGI gripper and set an Orbbec Femto Bolt on the front side of the workspace. The default image size of the captured RGB-D stream is $1280\times 960$. The setting is shown as Fig.~\ref{fig:ur5e}.

For the humanoid robot, we use the Unitree G1 equipped with Inspire RH56DFX dexterous hands and Intel RealSense D435i RGB-D camera. The image size for the RGB-D images is set as $640\times480$. The setting is shown as Fig.~\ref{fig:G1}. 

More experiments can be found in Fig.~\ref{fig:g1_exp} and the video in supplementary material.

\begin{figure}[h!]
\centering
\includegraphics[width=0.5\linewidth,trim={0cm 0cm 0cm 0cm}]{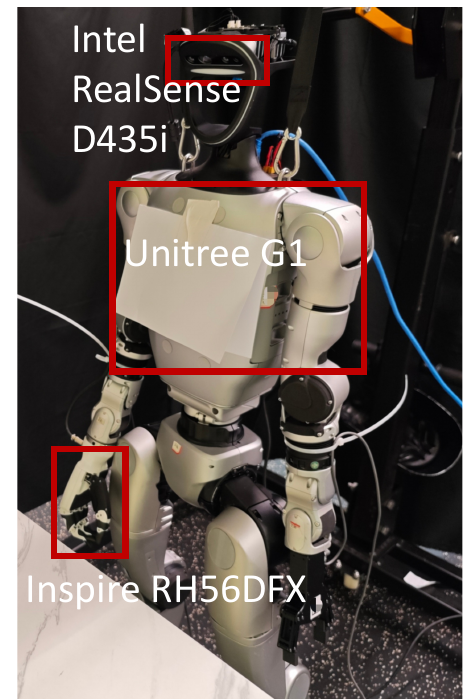}
\caption{The experiment setting of a humanoid robot with dexterous hands.} 
\label{fig:G1}
\end{figure}

\section{Failure Cases}
\begin{figure}[h!]
\centering
\includegraphics[width=1.0\linewidth,trim={0cm 0cm 0cm 0cm}]{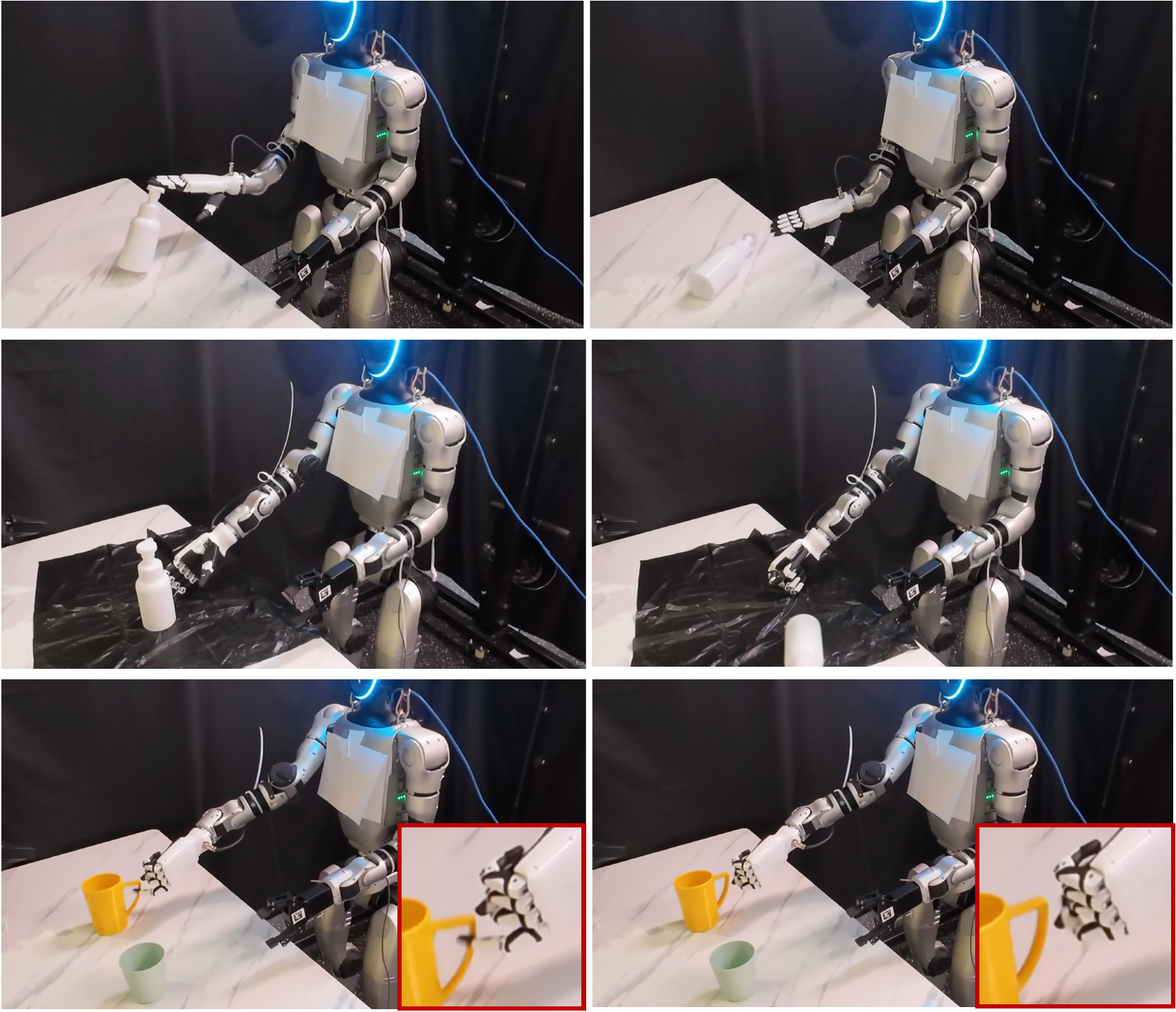}
\caption{\textbf{Failure cases.} Row 1 \& 2: Rough grasping motions lead to collisions and overturns. Row 3: The cumulative error in hand-eye calibration leads to inaccurate positions.} 
\label{fig:failure}
\end{figure}

\begin{figure*}[h!]
\centering
\includegraphics[width=0.98\linewidth,trim={0cm 0cm 0cm 0cm}]{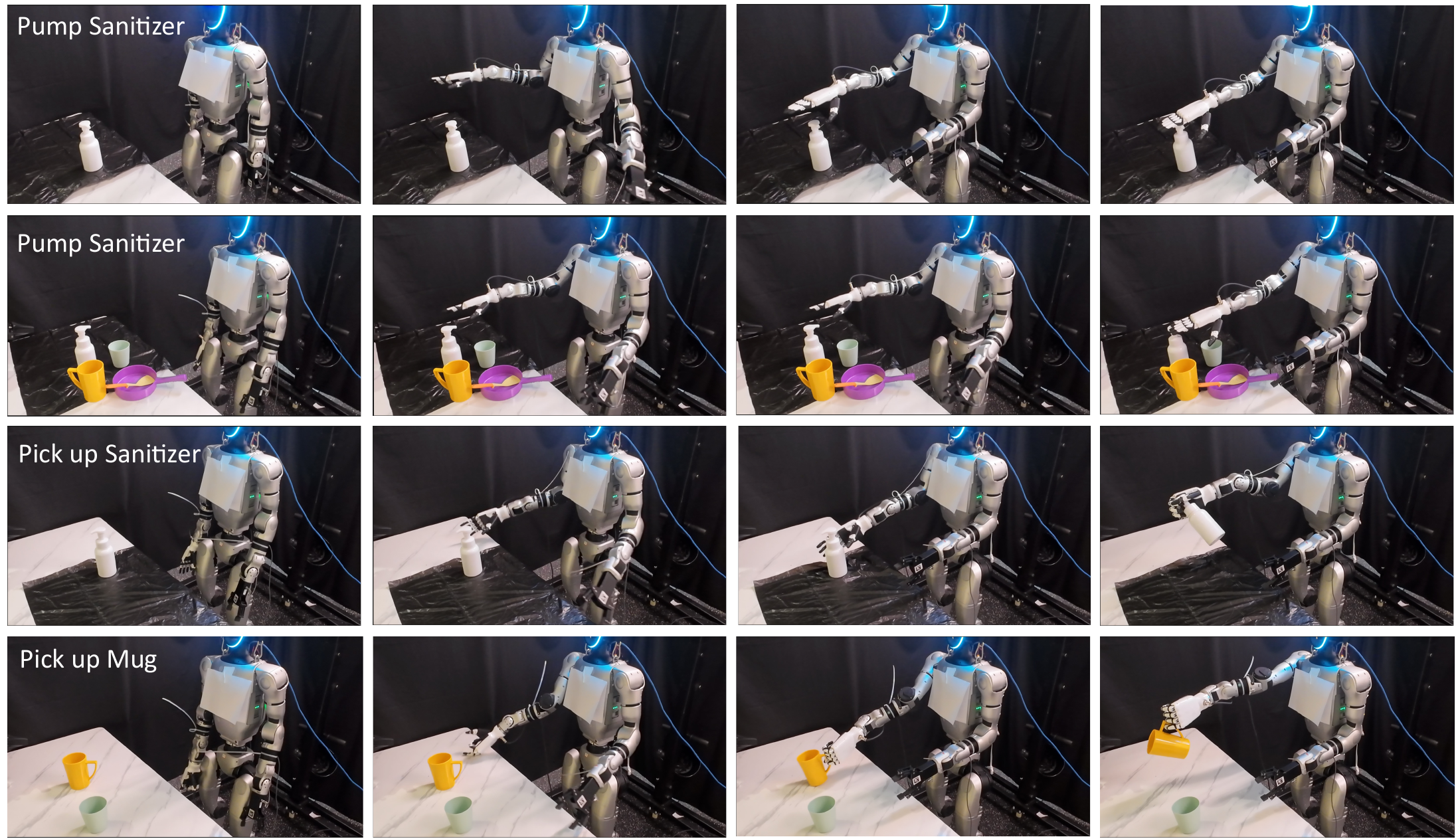}
\caption{Demonstrations of manipulation tasks completed by the Unitree G1. } 
\label{fig:g1_exp}
\end{figure*}

The failure cases in the real-robot experiments can be summarized as two main points: 
\begin{itemize}
    \item We use the Inverse Kinematics (IK) to calculate the motions based on the target pose, which leads to rough grasping motions. The lack of dexterous grasping policy leads to collisions and overturns, as shown in the first and second row of Fig.~\ref{fig:failure}. In the future, we aim to train ACT or Diffusion Policy for the grasping motion optimization.
    
    \item For the calibration of the relationship between the hand and eye of humanoid robots, our measurements may have certain errors, which could lead to the failure of tasks requiring fine operations, as shown in the third row of Fig.~\ref{fig:g1_exp}.
\end{itemize}

\section{Tracking Performance}
\begin{figure*}[h!]
\centering
\includegraphics[width=0.98\linewidth,trim={0cm 0cm 0cm 0cm}]{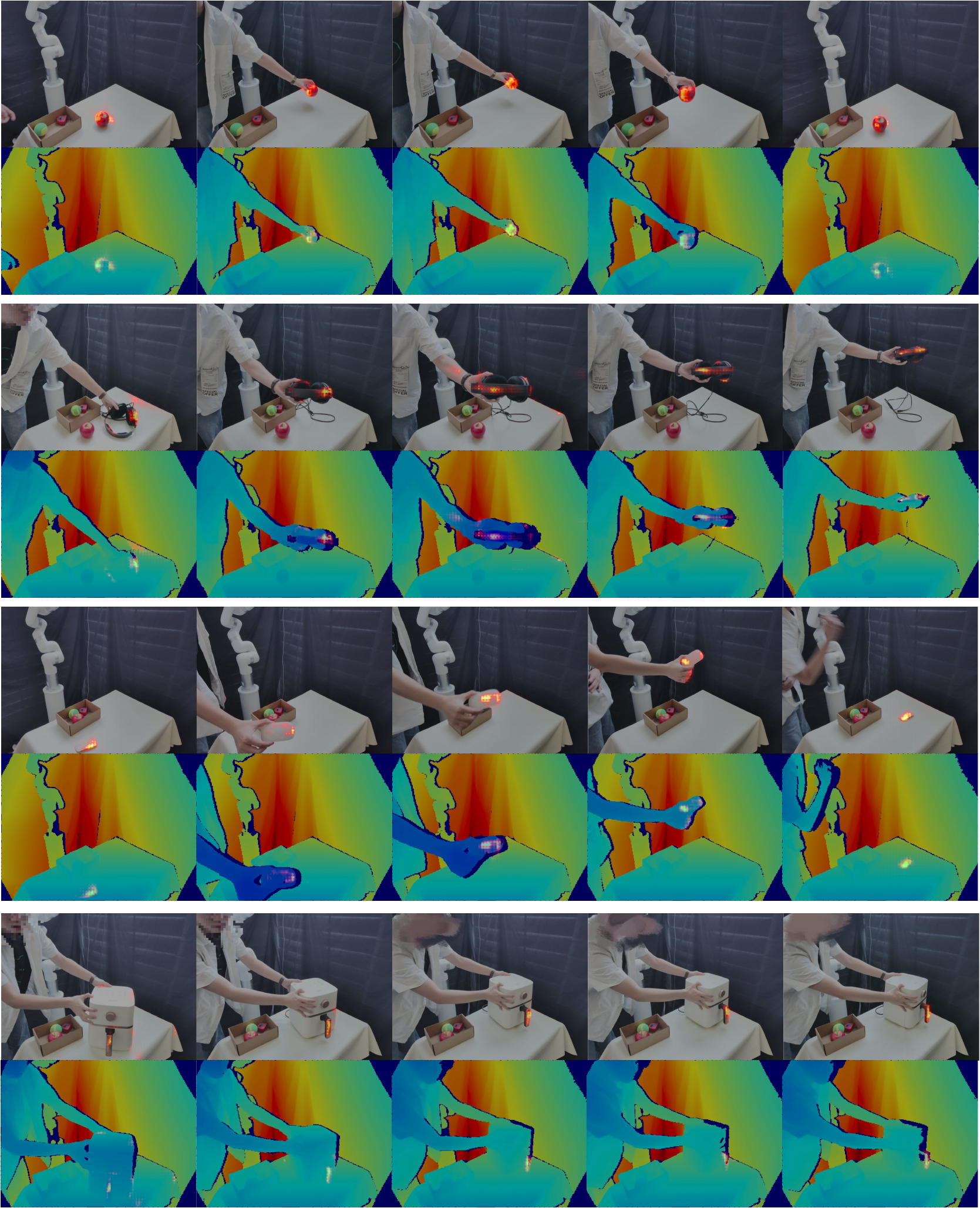}
\caption{\textbf{The visualization of tracking performance of \model.} We track the object \textit{apple, earphone, mouse, oven handle} from top to bottom.} 
\label{fig:track}
\end{figure*}

The real-time performance of our \model~makes tracking moving objects while reasoning affordance possible. Time cost can be found in Table 3 in the main manuscript. We show some real-time tracking $\&$ affordance reasoning examples in Fig.~\ref{fig:track}. The qualitative results demonstrate the fast, precise, and robust tracking performance of our \model.
More examples can be found in the video of supplementary material.




\end{document}